\documentclass[english,final,article]{jcaa}
\usepackage{multirow}

\title{Real-time Ultrasound-enhanced Multimodal Imaging of Tongue using 3D Printable Stabilizer System: A Deep Learning Approach}
\author[1]{M. Hamed Mozaffari\thanks{mmoza102@uottawa.ca}}
\author[1]{Won-Sook Lee}
\affil[1]{School of Electrical Engineering and Computer Science, University of Ottawa, Ottawa, Canada}

\begin{document}

\twocolumn[

\maketitle

\vspace{1em}
\hrule
\vspace{1em}

\begin{abstract}
Despite renewed awareness of the importance of articulation, it remains a challenge for instructors to handle the pronunciation needs of language learners. There are relatively scarce pedagogical tools for pronunciation teaching and learning. Unlike inefficient, traditional pronunciation instructions like listening and repeating, electronic visual feedback (EVF) systems such as ultrasound technology have been employed in new approaches. 
Recently, an ultrasound-enhanced multimodal method has been developed for visualizing tongue movements of a language learner overlaid on face-side of the speaker's head. That system was evaluated for several language courses via a blended learning paradigm at the university level. 
The result was asserted that visualizing articulator’s system as biofeedback to language learners will significantly improve articulation learning efficiency. 
In spite of that successful usage of multimodal technique for pronunciation training, it still requires manual works and human manipulation. In this article, we aim to contribute to this growing body of research by addressing difficulties of the previous approaches by proposing a new comprehensive, automatic, real-time multimodal pronunciation training system, benefits from powerful artificial intelligence techniques.
The main objective of this research was to combine the advantages of ultrasound technology, three-dimensional printing, and deep learning algorithms to enhance the performance of previous systems. Our preliminary pedagogical evaluation of the proposed system revealed a significant improvement in flexibility, control, robustness, and autonomy.
\end{abstract}


\begin{motclef}
Technologie à ultrasons pour la formation à la prononciation \sep
Production de la parole et visualisation de la langue par ultrasons \sep
Apprendre en profondeur pour la prononciation \sep
Extraction automatique du contour de la langue par ultrasons \sep
Approche multimodale améliorée par ultrasons \sep
Impression 3D pour la stabilisation de sonde à ultrasons. 
\end{motclef}

\hrule
\vspace{1.5em}
]

\saythanks

\section{Introduction and Previous Works}
Communication skill is one of the essential aspects of the second language (L2) acquisition so that it is often the first indication of a language learner’s linguistic abilities \cite{bird2018ultrasound}. Pronunciation directly affects many social interaction skills of a speaker, such as communicative competence, performance, and self-confidence. Previous studies revealed that other aspects of L2 learning, such as word learning, can be developed easier by accurate pronunciation \cite{johnson2018prior}. However, pronunciation learning is one of the most challenging skills to master for adult learners \cite{abel2015ultrasound} in traditional classroom settings. There is often no explicit pronunciation instruction for language learners because of limited time and lack of knowledge of effective pronunciation teaching and learning methods \cite{abel2015ultrasound}. 
\\ 
Standard practice for a language learner, outside of the class, is to imitate a native speaker’s utterances in front of a mirror limited to lip and jaw movements along with hearing of recorded acoustic data. In practice, it is difficult for an L2 learner to utter new words correctly without any visual feedback of a native speaker and lack of awareness of how sounds are being articulated \cite{abel2015ultrasound}, especially in cases where the target sounds are not easily visible \cite{bird2018ultrasound}. The positions and movements of the tongue, especially all but the most anterior part, cannot be seen in the traditional approach of listening and repeating word’s pronunciations \cite{bliss2017using}. Learners can only have proprioceptive feedback of their tongue location depends on practicing sounds (vowels, liquids, or others) and the amount of contact their tongue makes with the teeth, gums, and palate \cite{wilson2006ultrasound}. 
\\
Many previous investigations in the literature of L2 pronunciation acquisition and ability have been focused on acoustic studies dealing with the sound that is produced and infer the articulation that created the sound \cite{wilson2006ultrasound}. Nevertheless, both acoustic and articulatory studies are undoubtedly valuable tools for understanding the progress of an L2 learner. The latter one can often give a more accurate picture of the actions performed by the pronunciation learner while it looks directly at the articulators (e.g., the tongue, the lips, and the jaw) \cite{wilson2006ultrasound}. Employing acoustic data alone might jeopardize the understanding of L2 learners in mapping the acoustic information onto articulatory movements. Having seen the articulators directly by learners, they can probably improve their pronunciation by the perception of the articulatory adjustments \cite{wilson2006ultrasound}. Therefore, an effective way of pronunciation teaching and learning includes listening and repeating using both acoustic and visual articulatory information.
\\
Visual feedback approaches have been developed over the past decades to enable L2 learners to see moving speech articulators during a speech or a training session, benefiting from a range of tools called Electronic Visual Feedback (EVF) \cite{lambacher1999call} including ultrasound imaging, electromagnetic articulography (EMA), and electropalatography (EPG) \cite{bliss2018computer}. Among those technologies, ultrasound imaging is particularly non-invasive, safe, offer high dimensional continuous real-time data with acceptable frame-rate, portable, versatile, user-friendly, widely available, and increasingly affordable. Furthermore, ultrasound technology is capable of recording and illustrating the whole regions of the tongue (although the mandible sometimes obscures the tongue tip \cite{bird2018ultrasound}) during both dynamic and static movements. Other imaging modalities such as MRI and X-ray (more specifically cinefluorography) are also capable of showing a mid-sagittal view of the tongue. However, these techniques are often prohibitively expensive, non-accessible, and invasive \cite{davidson2006comparing}.
\subsection{Ultrasound Tongue Imaging}
\begin{figure}[ht]
\includegraphics[width=\columnwidth]{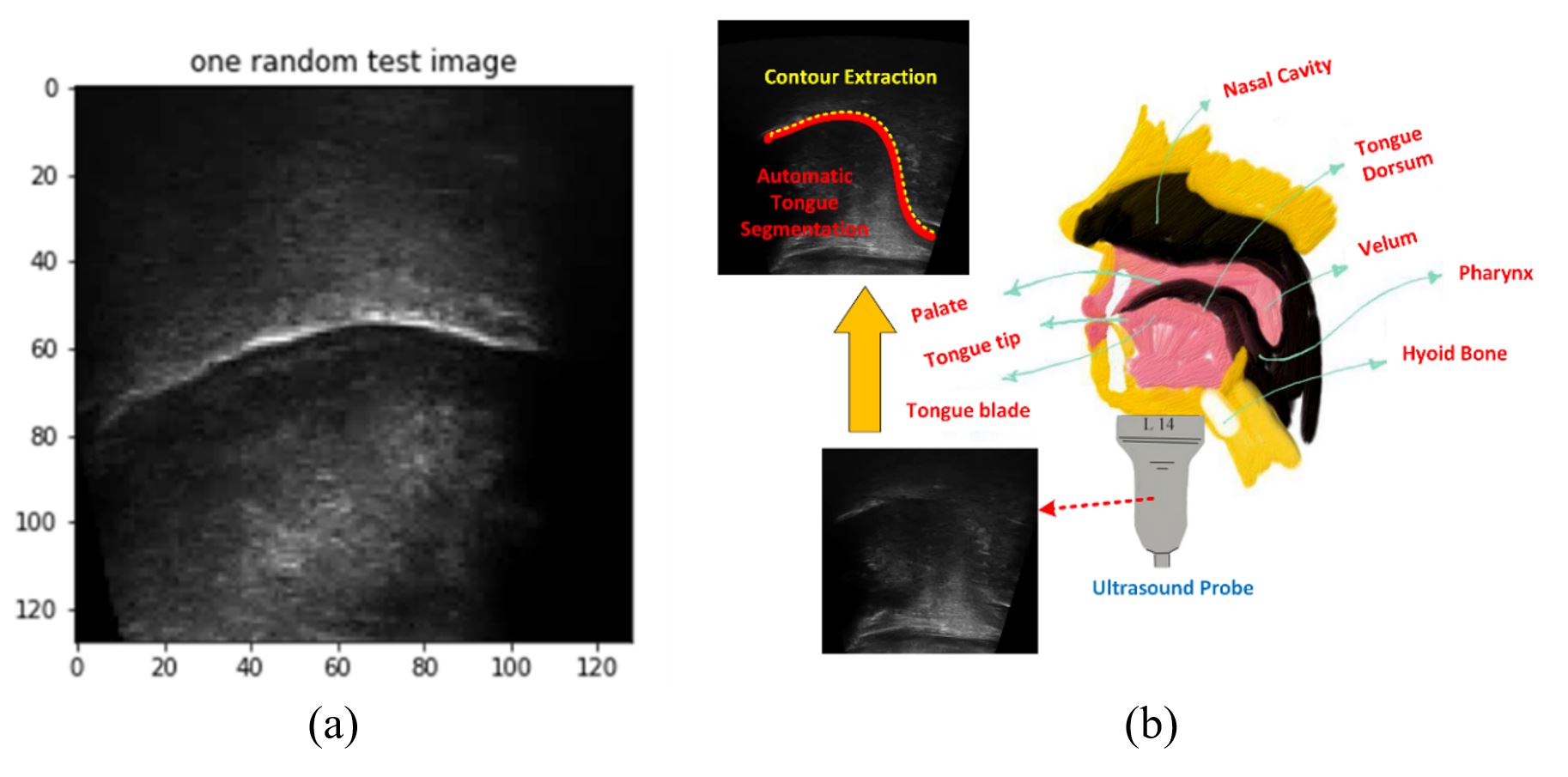}
\caption{\label{fig:FIG1}{Interpretation of Ultrasound video frames by non-expert L2 learners requires knowledge of tongue anatomy and understanding ultrasound data: a) one sample of ultrasound video frame after cropping into squared sized format (which side is the tip of the tongue?), b) Tongue contour can be highlighted for better understanding of the tongue gestures in video frames.}}
\end{figure}

About 40 years ago, one-dimensional ultrasound was first used effectively for illustration of one point at a time on the tongue’s surface \cite{kelsey1969ultrasonic}, while the two-dimensional ultrasound (B-mode settings for mid-sagittal or coronal view) has been employed in speech research since 25 years ago \cite{sonies1981ultrasonic}. Nevertheless, due to the recent development of ultrasound imaging technology with higher image quality and greater affordability, it became an essential tool for visualizing the speech articulators in speech research and for pedagogical use in the acquisition of L2 pronunciation \cite{bird2018ultrasound, wilson2006ultrasound}. 
\\
Ultra-high frequency sound both emitted and received by piezoelectric crystals of ultrasound transducer/probe creates echo patterns which are decoded as an ultrasound image. Ultrasound signal penetrates and traverses linearly through materials with uniform density but reflects from dense substances such as bone. With the ultrasound transducer held under the chin and with the crystal array lying in the mid-sagittal plane of the head, the ultrasound screen displays information about the superior surface of the tongue from the tongue root to near the tip along the mid-sagittal plane \cite{campbell2010spatial} (see Figure~\ref{fig:FIG1}.b). Procedures and techniques of ultrasound image reconstruction and acquisition, specifically for tongue imaging have been comprehensively enumerated and explained in \cite{stone2005guide}.
\\
One particular linguistically valuable property of ultrasound imaging is the capability of simultaneous visualization of the front and back of the tongue \cite{wilson2006ultrasound}. For instance, in the production of some consonant such as /l/ and /r/, the time intervals between gestures of different tongue’s regions (e.g., the tongue tip, tongue dorsum, and tongue root) are essential, and they depend on the position of the consonant in the syllable (i.e., onset versus coda) \cite{tateishi2013does, campbell2004gestural, campbell2010spatial}. Therefore, an L2 learner can use ultrasound imaging to understand the source of those mistiming errors, results in more accurate pronunciation. 
\\
The efficacy of using ultrasound imaging on the pronunciation of North American /r/ and /l/ phonemes has been proven by depicting the complexity of the tongue’s shape for L2 language learners \cite{adler2007use, wilson2006ultrasound}. Although ultrasound has been utilized successfully in L2 pronunciation training, interpretation of raw ultrasound data for language learners, especially in a real-time video stream, is a challenging task. Moreover, to understand ultrasound videos accurately, a language learner ought to know the tongue's gestures and structures as well as interpretation of ultrasound data \cite{bliss2016ultrasound}. Figure~\ref{fig:FIG1}.a shows a sample of ultrasound tongue frame.

\subsection{Ultrasound-enhanced Multimodal Approach}
\begin{figure}[ht!]
\includegraphics[width=\columnwidth]{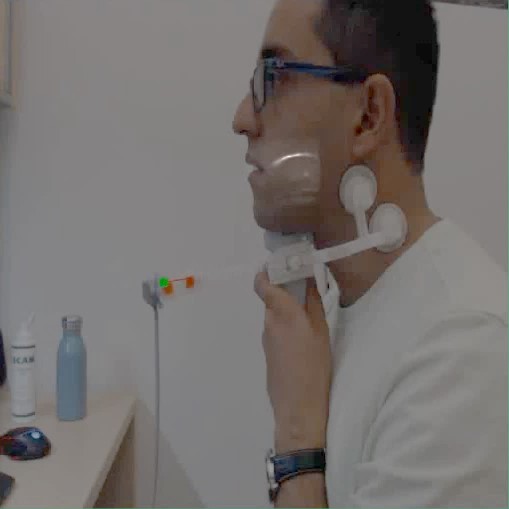}
\caption{\label{fig:FIG2}{L2 learners are able to see their tongue gestures during a pronunciation training session using an ultrasound-enhanced multimodal approach. Real-time perceiving the tongue location in the mouth is significantly easier and more effective than looking at multimodal recorded videos.}}
\end{figure}

An ultrasound-enhanced multimodal approach is a novel alternative for assisting L2 pronunciation learners in understanding and perceiving the location of their tongue on ultrasound video frames. Figure~\ref{fig:FIG2} illustrates one frame captured using our ultrasound-enhanced multimodal technique. Significant pioneer studies of this method have been accomplished by researchers in linguistics department at the University of British Columbia (UBC) \cite{abel2015ultrasound, bird2018ultrasound, bliss2018seeing, bliss2016ultrasound, gick2008ultrasound, wilson2006ultrasound, yamane2015ultrasound}. The key technological innovation of those systems is the use of mid-sagittal ultrasound video frames of the tongue, manually overlaid on the external profile views of a speaker’s head to allow learners to observe speech production on their face profile. In order to highlight the tongue region in ultrasound frames, pixels' intensities related to the tongue region was manipulated manually by pink colour \cite{gick2008ultrasound}.
Benefits of a multimodal approach for pronunciation language training have recently been investigated \cite{bliss2018computer, mozaffari2019grapp, mozaffari2018guided}. However, manual work is still extensive in many stages (pre-processing such as image enhancement, during the exam like overlaying ultrasound frames on RGB video frames, and post-processing including highlighting of the tongue region and audio/video synchronization). Furthermore, the overlaid videos come with some non-accuracy due to the lack of transformational specification (exact scale, orientation, and position information) of the ultrasound frame for superimposing on face profile. Moreover, exact transformations cannot be generalized for face profile of any language learner, often restricted to one position of the head. Therefore, the head of a language learner should be stabled during training session. Accurate synchronization between ultrasound data, video frames, and acoustic records is another challenge for previous studies. Beside those difficulties, quantitative study of tongue movement only viable after freezing a target frame or during post-processing of recorded frames \cite{abel2015ultrasound, bernhardt2005ultrasound}. 

\subsection{Artificial Intelligence for Second Language Acquisition}
Artificial intelligence (AI) is a branch of computer science when machines can do tasks that typically require human intelligence \cite{hamet2017artificial}. Machine learning is a subset of AI, where machines can learn by experiencing and acquiring skills without human involvement. Inspired by the functionality of the human brain, artificial neural networks learn from large amounts of data to perform a task repeatedly. Deep learning algorithms are artificial neural networks with various (deep) layers similar to human brain structure \cite{lecun2015deep}. Deep learning-based methods and their applications in image processing literature such as object detection and image segmentation have been a research hotspot in recent years. Deep learning methods are powerful in automatic learning of a new task, while unlike traditional image processing methods, they are capable of dealing with many challenges such as object occlusion, transformation variant, and background artifacts \cite{chen2017deeplab, guo2016deep, zhao2019object}.
\\
Recent technology-assisted language learning methods, such as multimodal approaches using ultrasound imaging, have been successfully employed for language pronunciation teaching and training, providing visual feedback of learner’s whole tongue movements and gestures \cite{abel2015ultrasound, antolik2019effectiveness, bernhardt2005ultrasound, bird2018ultrasound, gick2008ultrasound, hueber2013ultraspeech, yamane2015ultrasound}. However, this technology is still far from commercializing for use in all language training institutes. The authors observe several main gaps in the current literature: 
\begin{itemize}
\item Ultrasound-enhanced multimodal methods require manual pre-processing and post-processing works, including image enhancement, freezing for further analysis, and superimposing of ultrasound frames and side-face profile video frames. Moreover, manual synchronizing is required between ultrasound frames, video frames, and audio data. All manual works are time consuming, subjective, and error-prone task, which require a knowledge of video and audio editing toolboxes \cite{abel2015ultrasound, bird2018ultrasound}. The performance quality of each study is evaluated only by qualitative investigation, and quantitative study of the method is only limited to post-processing stages, while it requires freezing target frames following cumbersome manual works.  
\item It is a challenging task for a non-expert language learner to interpret ultrasound data in real-time \cite{bliss2016ultrasound} without having any knowledge of tongue structure and ultrasound imaging. Colouring the tongue regions in ultrasound data \cite{bird2018ultrasound} cannot be an efficient and generalized approach for this problem due to the different ultrasound image data characteristics as well as the requirements of additional manual work. 
\item Our experimental study revealed that language learners could understand the gestures of their tongue better in real-time using ultrasound-enhanced multimodal visualization approach than previous recorded offline systems. Instead of using a guideline on the screen (usually on the palate \cite{bernhardt2005ultrasound}), using an automatic tracking technique, a language learner can perceive the real-time location of the tongue respect to landmarks of the face. 
\item Due to the dependency of previous ultrasound multimodal biofeedback methods on the subject’s face specifications \cite{mozaffari2019grapp}, those studies are not applicable for other medical applications such as ultrasound augmented reality, and manual works for primary system’s modifications are inescapable. 
\item Previous stabilization methods have been designed only for one specific type of ultrasound probe without any flexibility for user’s head position. To overlay ultrasound video frames on the user's face, the head should be stabled during recording which makes the training session non-conformable for the user. Furthermore, different ultrasound helmets and probe stabilizers are not accessible in any research and teaching departments. 
\end{itemize}
The motivation of this paper is to provide a real-time and fully automatic toolkit to address those difficulties by utilizing state of the art automation benefits from deep learning techniques and a novel 3D printable stabilizer device named "UltraChin". Our pronunciation training system comprises of several modules that can be utilized efficiently for L2 pronunciation teaching and learning without difficulties of previous systems. Using the UltraChin, users can hold the probe without concern about occlusion and restriction for head movements.
\\
The rest of this article is structured as follows. Section~\ref{section:SEC2} describes our proposed automatic and real-time ultrasound-enhanced multimodal pronunciation system. In this section, we explained each module of our system separately in details. Section~\ref{section:SEC3} summarizes our experimental results and studies. Finally, section \ref{section:SEC4} discusses and concludes our paper as well as potential future directions.

\section{Methodology}
\label{section:SEC2}

Our fully automatic, real-time, and subject independence ultrasound-enhaced multimodal pronunciation training system comprises of several modules. In the following subsections, we explain each module separately in details. 

\subsection{UltraChin for Stabilization and Tracking}

\begin{figure}[ht]
\includegraphics[width=\columnwidth]{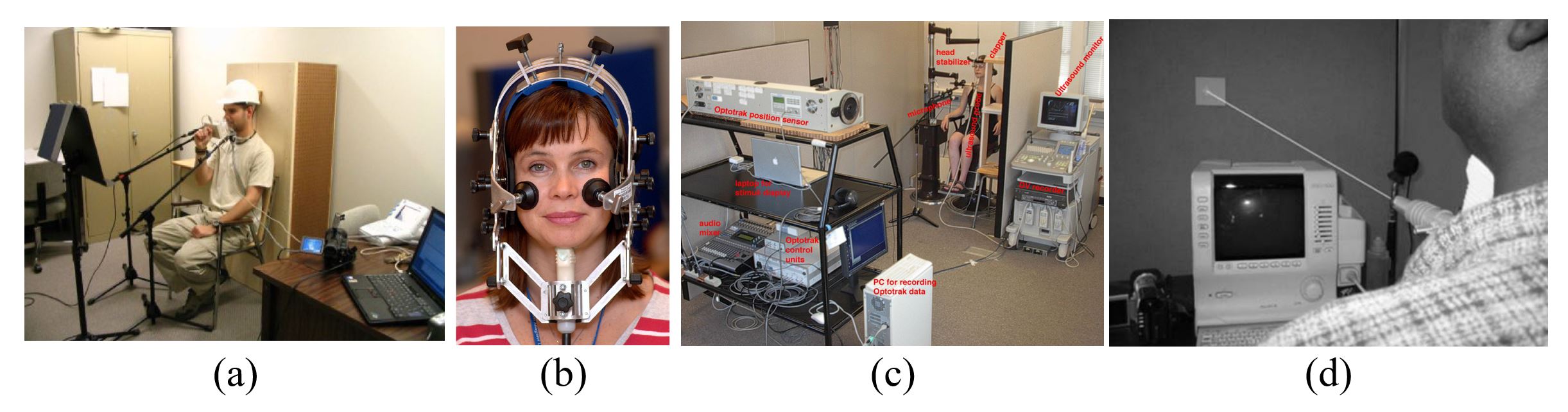}
\caption{\label{fig:FIG3}{An illustration of some stabilization methods for head and probe. a) Helmet fixed to the wall \cite{menard2011development}, b) Designed helmet keep the probe under the chin \cite{scobbie2008looking}, c) Optical tracking system for the head alignment \cite{campbell2010spatial}, and d) A laser connected \cite{gick2002use} to the probe detects any movements of the neck and the probe slippage.}}
\end{figure}

In previous ultrasound-enhanced multimodal studies, head of a language learner should be stabilized during video and ultrasound data collection as well as superimposing video frames \cite{abel2015ultrasound, bird2018ultrasound}. This obligation is to ensure that ultrasound videos are controlled with respect to the orientation of the probe as well as providing a positional reference in quantitative assessments \cite{bird2018ultrasound}. A consequence of this restriction is considerable reducing flexibility of the user's head movements. Moreover, the language learner should concentrate on the user interface for a long time with those limited movements where it might result in body fatigue and eye strain, ultimately reducing the effectiveness of the L2 training session (see Figure \ref{fig:FIG3} for some samples of stabilization methods). 

\begin{figure}[ht]
\includegraphics[width=\columnwidth]{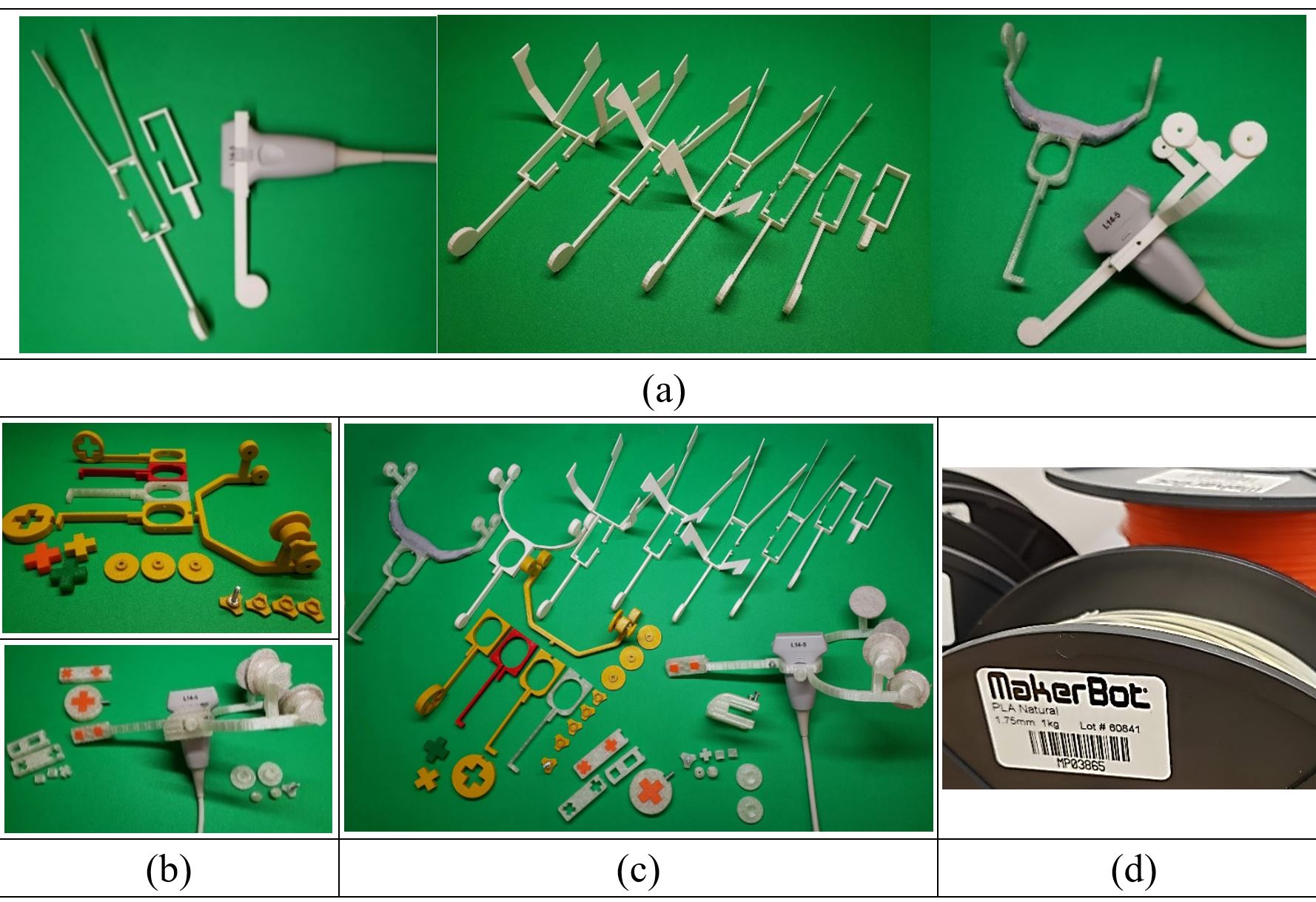}
\caption{\label{fig:FIG4}{UltraChin: a) Integrated designs, b) Modular designs, c) Different versions and parts, d) Natural PLA materials for printing.}}
\end{figure}

\begin{figure}[ht]
\includegraphics[width=\columnwidth]{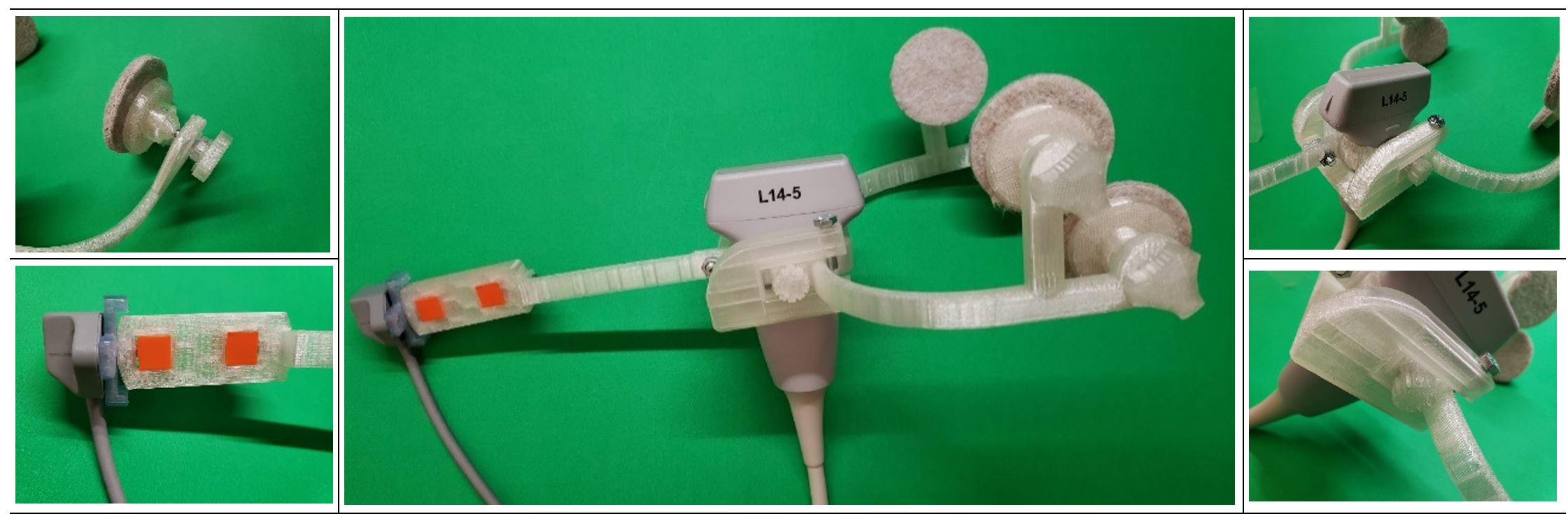}
\caption{\label{fig:FIG5}{The last generation of UltraChin comprises of several modules. This version is universal and usable for different ultrasound probes as well as capable of being attached to other sensors.}}
\end{figure}

In a recent study \cite{mozaffari2018guided, mozaffari2019grapp}, the head has a flexibility without any stabilization due to the automatic face tracking. However, in this approach, different face profile (different genders, skin characteristics, and ages) might provide a different result in each experiment, and ultrasound video is overlapped on the face with some non-accuracy. The freehand method can not guarantee that the probe orientation is always in the mid-sagittal plane during a speech. In order to make our system independent from the user’s face profile, we designed UltraChin which is a universal 3D printable device compatible for any type of ultrasound probes. 
\\
In addition to the aims aforementioned above, UltraChin is used for two other purposes in our system: I) As a reference marker for probe tracking module, II) For keeping the probe under the chin aligned with the mid-sagittal plane of L2 learner's face. UltraChin provides reference points for transformations, which are required to overlay ultrasound video frames on the user’s face without any limitation for head movements. 
\\
UltraChin was created after several generations of designing, 3D printing, and testing on language learners (see figure \ref{fig:FIG4} for several generations of UltraChin). In order to make each part of the UltraChin, we used SolidWorks software. In the last generation (see figure \ref{fig:FIG5}), we used natural materials in the process of 3D printing for sensibility prevention due to the contact of human skin with plastic which was not considered in previous similar devices \cite{scobbie2008head, spreafico2018ultrafit}. Furthermore, adding an extra part, users can attach other types of sensors such as electromagnetic tracking sensor. 

Unlike the previous helmets and stabilizers devices \cite{derrick2018three, scobbie2008head, spreafico2018ultrafit} for ultrasound tongue imaging, UltraChin is a universal device capable of connecting to different types of ultrasound probe. It is fully printable and publicly available capable of attaching to other sensors. One unique characteristic of UltraChin is that the ultrasound probe is held by the user which makes the process of data acquisition more comfortable and even more accurate after training the user to keep the probe correctly. This freehand ability enables users to utterance words with more comfort while the optimized pressure of the probe on the chin is set by the user after several training sessions, results in better image quality and less slippage of the probe. Printable designs of the UltraChin can be find freely in the internet \footnote{https://github.com/HamedMozaffari/UltraChinDesigns}.

\subsection{ProbeNet for Probe Tracking and Video Overlaying}
\begin{figure*}[ht]
\includegraphics[width=\textwidth]{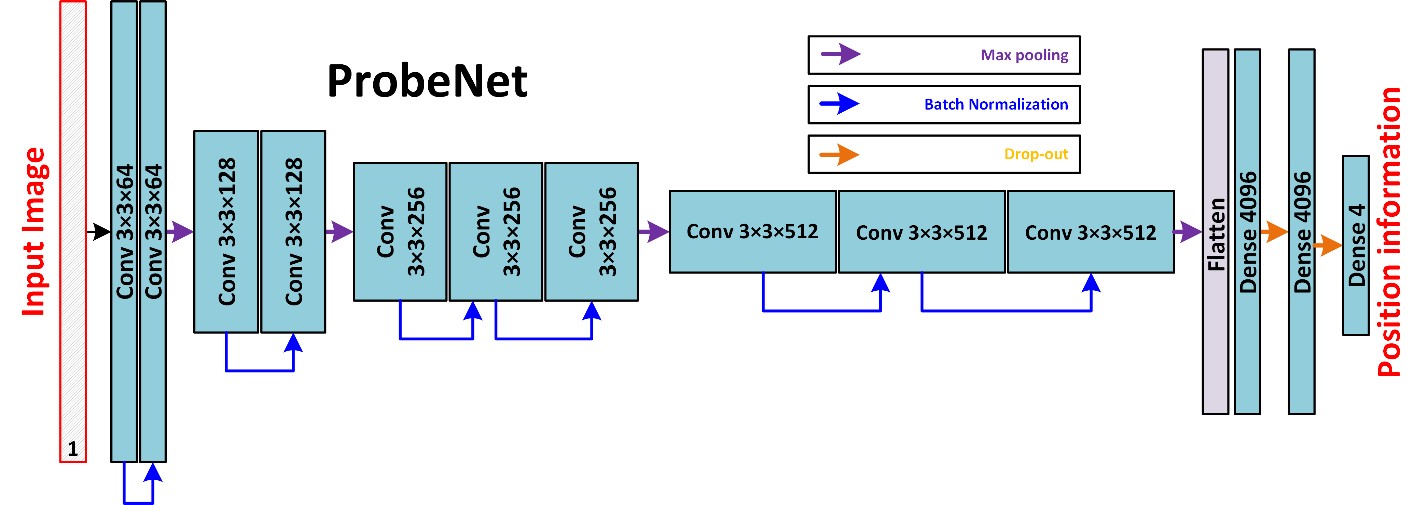}
\caption{\label{fig:FIG6}{ProbeNet architecture for tracking landmarks on the UltraChin. Output layer provides two sets of numbers as vertical and horizontal positions of each landmark.}}
\end{figure*}

Object localization (and/or detection) is determining of where objects are located in a given image using bounding boxes encompass the target object. Various deep learning methods have been proposed for object localization in recent years \cite{yilmaz2006object, zhao2019object}. Similar to facial landmark detection \cite{wu2019facial} when several face’s key points are detected as landmarks, we defined two key points on UltraChin as markers for the sake of probe tracking. Two landmarks (key points) are two orange cubes which we considered two upper-left corners (see two embedded orange squares in figure \ref{fig:FIG5}).
\\
The two markers are tracked automatically in real-time using our new deep convolutional network (we named that ProbeNet). In this method, positions of the two key points on UltraChin provide us transformational information in each frame, comprises of probe orientation, location, and a reference for scaling of the ultrasound data. We designed ProbeNet specifically for the probe tracking problem by inspiring from VGG16 architecture \cite{simonyan2014very}. 
\\ 
Tracking of an ultrasound probe has been already accomplished using different kinds of devices such as electromagnetic, optical, GPS, and mechanical sensors \cite{mozaffari2017freehand}. However, the primary motivation of those studies is to track the probe in three-dimensional space (usually with 6 degrees of freedom (DOF)). In this study, we considered a simplifying assumption: \textit{ultrasound probe and language learner’s face are moving parallel respect to the camera lens both in two-dimensional planes} (see Figure \ref{fig:FIG2} where ultrasound frame, segmented dorsum with white color, and two orange markers are parallel respect to the camera). Under this assumption, tracking of the probe only requires the calculation of the location in a two-dimensional plane instead of three-dimensional space. For this reason, we selected two key points on the UltraChin, and the tracking problem was converted from three-dimensional space to two-dimensional space. 
\\
Figure \ref{fig:FIG6} illustrates the detailed architecture of the ProbeNet for marker detection and tracking. ProbeNet comprises of several standard convolutional layers followed by ReLU activation function and batch-normalization for more efficient training. In the last block, we used a dense layer with four neurons, which provides 2D positions of upper-left side of the two markers as well as drop-out of 50 percent for better generalization over our annotated dataset. During a pronunciation training session, positional information of the two markers are predicted by ProbeNet. These data are used to find the best position, orientation, and scaling of ultrasound frame overlaying on L2 leaner's face. It is noteworthy to mention that a specific calibration procedure is required to convert positional information into orientation and scaling data in visualization module.

We consider a ratio of the distance between the two markers and width of the ultrasound probe for ultrasound frame scaling. Triangular calculations between the two markers provides ultrasound frame translation and orientation over the user's face. 

\subsection{Tongue contour segmentation, extraction, and tracking}
\begin{figure*}[ht]
\includegraphics[width=\textwidth]{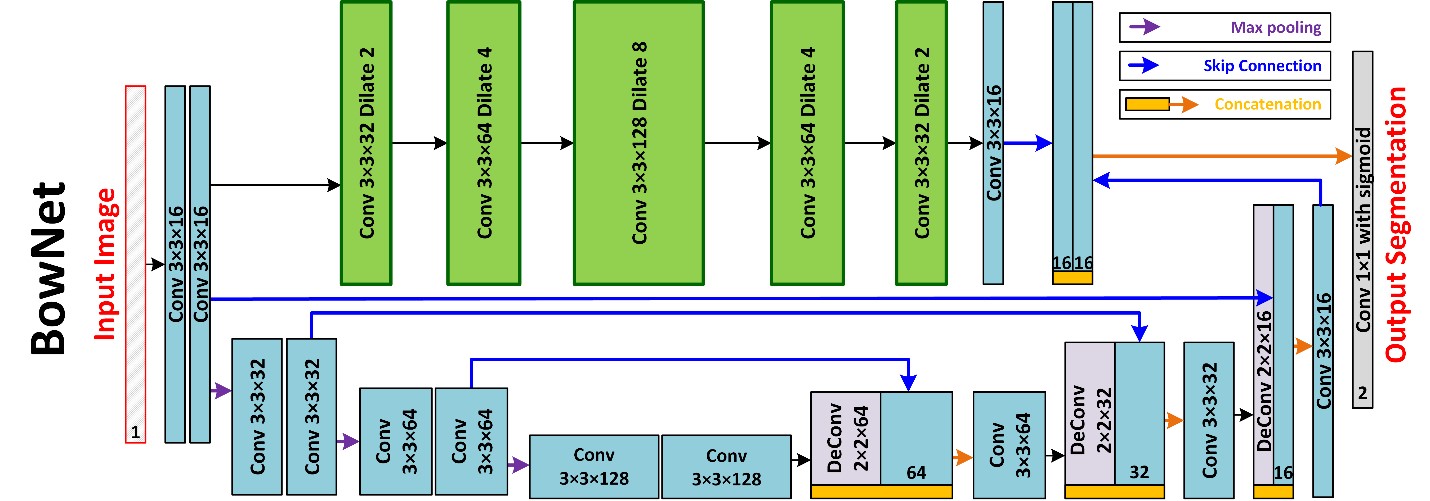}
\caption{\label{fig:FIG7}{BowNet network architecture for tongue contour extraction automatically and in real-time \cite{mozaffari2019bownet}}}
\end{figure*}

Typically, during tongue data acquisition, ultrasound probe beneath the user’s chin images tongue surface in mid-sagittal or coronal view \cite{slud2002principal} in real-time. Mid-sagittal view of the tongue in ultrasound data is usually adapted instead of coronal view for illustration of tongue region, as it displays relative backness, height, and the slope of various areas of the tongue. Tongue dorsum can be seen in this view as a thick, long, bright, and continues region due to the tissue-air reflection of ultrasound signal by the air around the tongue (see Figure \ref{fig:FIG1}). 
\\
This thick white bright region is irrelevant, and the tongue surface is the gradient from white to the black area at the lower edge \cite{stone2005guide}. Although the tongue contour region can be seen in ultrasound data, there are no hard structure references. For this reason, it is a challenging task to locate the tongue position and interpret its gestures without any training \cite{stone2005guide}. Furthermore, due to the noise characteristic of ultrasound images with a low-contrast property, it is an even more laborious task for non-expert users to follow the tongue surface movements, especially in real-time applications \cite{bliss2016ultrasound}.
\\ 
Few previous multimodal studies proposed manual highlighting of the tongue region with different colour, which is not applicable for automatic and real-time applications. In this work, we utilized a fully-automatic and real-time image segmentation method from deep learning literature called BowNet \cite{mozaffari2019bownet} to track the surface of the tongue in video frames (as a continuous high-lighted thick region). A detailed description of the BowNet architecture is beyond the scope of the present paper, and we described only several critical aspects of that model briefly for the sake of presentation.
\\
Benefiting from several innovative techniques such as deconvolutional, dilated and fully convolutional layers, parallel paths, and skip connections, BowNet model could reach to higher accuracy with a robust performance in the problem of ultrasound tongue contour extraction in comparison to similar methods \cite{laporte2018multi}. At the same time, there was no compromising for other aspects of the BowNet model like computational cost, the number of trainable parameters, or real-time performance \cite{mozaffari2019bownet}. Figure \ref{fig:FIG7} presents the network structure and structural connections between different layers of the BowNet model \cite{mozaffari2019bownet}. 

Enhancing ultrasound frames by highlighting the tongue dorsum region enables language learners to focus on handling challenges of pronunciation learning instead of the interpretation of ultrasound data. Furthermore, extracted tongue contours provide teachers and language researchers valuable information for quantitatively comparison studies. It is noteworthy to mention that tongue contour extraction is done after tongue region segmentation using an image processing technique such as skeletonizing or just keeping the top pixels of the tongue region \cite{mozaffari2018guided, zhang1984fast, mozaffari2019grapp} (see Figure \ref{fig:FIG1}.b for a sample of tongue surface region and extracted contour). 

\subsection{Automatic and Real-time Pronunciation Training System}

\begin{figure*}[ht]
\includegraphics[width=\textwidth]{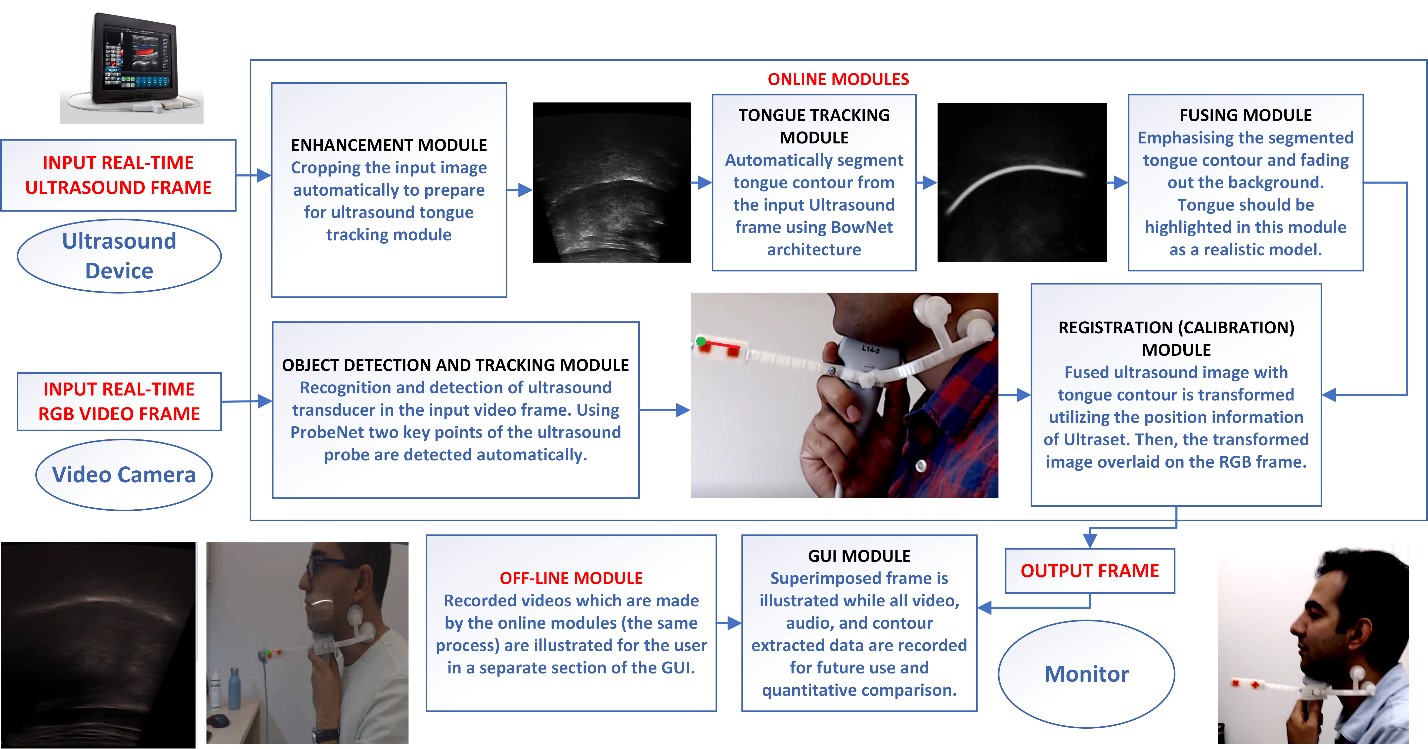}
\caption{\label{fig:FIG8}{The detailed architecture of our multimodal, real-time, and automatic pronunciation training system comprises of two main online and offline modules.}}
\end{figure*}

We deployed our pronunciation training system using Python programming language and several standard public libraries as a modular system to enable other researchers to improve or customize each module for any future research. Figure \ref{fig:FIG8} represents a schematic of our pronunciation training modules and their connections. As can be seen in the figure, there are two streams of data recording in our system, an off-line module which is used for recording videos by native speakers for teaching and an online module for L2 learners which is used for pronunciation training in real-time. 

Ultrasound data acquisition and analysis generally involves capturing both acoustic and ultrasound video frames together so that the audio track can help with identifying target sounds on the ultrasound video stream. Synchronization of those two data can be done during recording or can be part of post-processing which is an integral and challenging part of having an accurate analysis. Our multimodal system encompasses different data and techniques during its performance, including tracking of UltraChin's markers, ultrasound data stream visualization, tongue surface segmentation, tongue contour extraction, audio recording and playback, calibration and superimposing video frames, learner’s lips visualization (front side), and network connections between ultrasound and workstation. In order to have an approximately synced system, all these stages work together as the following procedure (see Figure \ref{fig:FIG8} for more details):

\begin{enumerate}
\item The data stream from RGB camera (Video and Audio) are captured and visualized in real-time. We used a Logitech Webcam with framerate of 30 fps connected to our workstation (a personal computer with a CPU of 7 cores and 16 GB of memory equipped with a GPU of NVidia GTX1080). Video and audio are already synced in this stage. 
\item Ultrasound stream video data is acquired and sent to the same workstation using Microsoft Windows remote desktop software. We employed a linear ultrasound transducer L14-38 connected to an Ultrasonix Tablet with settings of the tongue (depth of 7 cm, a frame rate of 30 fps \cite{stone2005guide}). It is noteworthy to explain that instead of working on ultrasound stream in our Python codes, we used a Window capturing library to grab ultrasound video from remote desktop software. This method enabled our system to be an ultrasound device independent where it can work with different ultrasound devices.
\item The current RGB Video frame is fed to our probe tracking module. Pre-trained ProbeNet model provides locations of two markers on the UltraChin (see the green dot (first marker) in Figure \ref{fig:FIG8} and the connected red line). In a predefined automatic calibration process, position, orientation, and probe head length are determined, and then they are sent to the visualization module. Simultaneously, the current ultrasound video frame is cropped, scaled, and fed to the BowNet model for the sake of tongue region segmentation. 
\item Results of previous stages, which are three images including cropped ultrasound frame, segmented tongue region, and RGB video frame, are superimposed using transformation information from ProbeNet and calibration calculations. 
\item Superimposed video stream is made by weighting the transparency of three frame data. The result is sent to the visualization module for illustration and recording. In this module, a designed graphical user interface (GUI) in Python language would illustrate several video streams from recorded datasets, real-time monitoring, audio data analysis bar, superimposed video, individual frames from ultrasound and video camera, results of real-time quantitative analysis, and possibly data from another ultrasound stream. There is also another camera for recording lip movements in front-view during a pronunciation training session. Development of this module is still in the early stages. 
\end{enumerate}

In our current system, a pronunciation learner or teacher can see several Windows in real-time separately on a display screen at the same time depends on the session target. For instance, as a speech investigation session, a researcher can see weighted ultrasound data superimposed on the face side in real-time, as well as separate non-overlaid ultrasound and RGB videos, accompany with prediction frame. Separated data will assist a researcher to compare different tongue contours qualitatively and quantitatively with a native speaker or recorded videos from previous sessions as a follow-up study in diagnosis of speech disorder. 
\\
Due to independence characteristic of our system respect to the number of image processing streams as a multimodal system, in a different scenario, using two ultrasound devices, L2 language teachers and learners can see and compare their tongues in real-time. Moreover, our system is capable of illustrating the difference between their tongue contours automatically. Due to the lack of the second ultrasound, we used recorded videos as the second reference video for our comparison studies and for capturing critical moments in the articulation (e.g., the stop closure in a stop articulation). 

\section{Experiments and Results}
\label{section:SEC3}
In this study, we proposed several modules, including two different deep learning models. The effectiveness of our system was evaluated for L2 language pronunciation teaching and learning in several aspects. In this section, we focus more on the linguistic perspective performance of our system as well as performance illustration of ProbNet, BowNet, and UltraChin.

\subsection{Probe Detection Module}
In order to train ProbeNet for tracking the markers on the UltraChin, we created a dataset comprises of 600 images of 3 different participants. During the exam session, the probe was kept by the user under the chin using UltraChin in the mid-sagittal plane. Recorded frame was annotated manually by placing two pre-defined key points on the upper-left side of orange markers on each frame. Dataset was divided into 80$\%$ training, 10$\%$ validation, and 10$\%$ testing sets. Finally, using our data augmentation toolbox (applying rotation, scaling, translation, and channel shift for images and the two key points), we created a dataset of 5000 images and their corresponding annotation information. 
\\
Adam optimization algorithm, with the first and second momentum of 0.9 and 0.999, respectively, was used to optimize Mean Absolute Error (MAE) loss function during training and validation. A variable learning rate with exponential decay rate and initial value of 0.001 was selected for the training of the ProbeNet. We trained the ProbeNet model for ten epochs (each with 1000 iterations) with mini-batches of 10 images. Our experimental results revealed the power and robustness of the ProbeNet in landmark detection and tracking task on UltraChin device. We got average MAE of 0.027 $\pm$ 0.0063 for ten times running of the ProbeNet on the test dataset. 

\subsection{Tongue Contour Extraction Module}
\begin{figure*}[ht!]
\includegraphics[width=\textwidth]{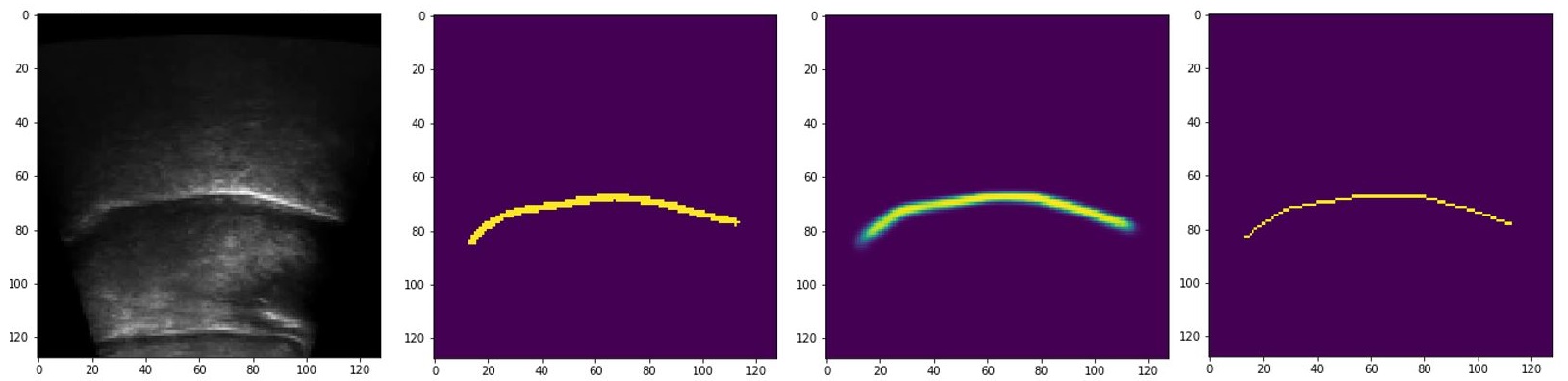}
\caption{\label{fig:FIG9}{One test sample used for testing the BowNet network. From left to right columns are ultrasound image, ground truth image, predicted map, extracted contour from the predicted map.}}
\end{figure*}

Few previous studies have used deep learning methods for tongue contour extraction with acceptable results \cite{laporte2018multi, zhu2019cnn}. For ultrasound contour tracking module, we used the most recent deep learning model in ultrasound tongue literature called BowNet. Figure \ref{fig:FIG7} represents the detailed architecture of the BowNet. For training settings of the BowNet, we followed the procedure in \cite{mozaffari2019bownet} for data from our ultrasound machine. Similar to the ProbeNet, for the training of the BowNet model, we separated the dataset into 80$\%$ training, 10$\%$ validation, and 10$\%$ test sets. 
\\
The BowNet model was trained and validated using online augmentation, and then it was tested separately on the test dataset. Figure \ref{fig:FIG9} presents a sample result of the BowNet model. Due to the more generalization ability of BowNet network, it can provide instances from different ultrasound machines with less false predictions. For more details about performance evaluation of the BowNet, refer to the original study \cite{mozaffari2019bownet}. 

The BowNet model was trained to work on data recorded from two ultrasound datasets. The tongue contour tracking performance of our system might be dropped for a new ultrasound data. A recent study in tongue contour extraction literature \cite{mozaffari2019transfer} has investigated the usage of domain adaptation for several different ultrasound datasets which can alleviate this difficulty significantly. 

\subsection{Accuracy Assessment of the UltraChin}

\begin{figure*}[ht]
\centering
\includegraphics[width=0.8\textwidth]{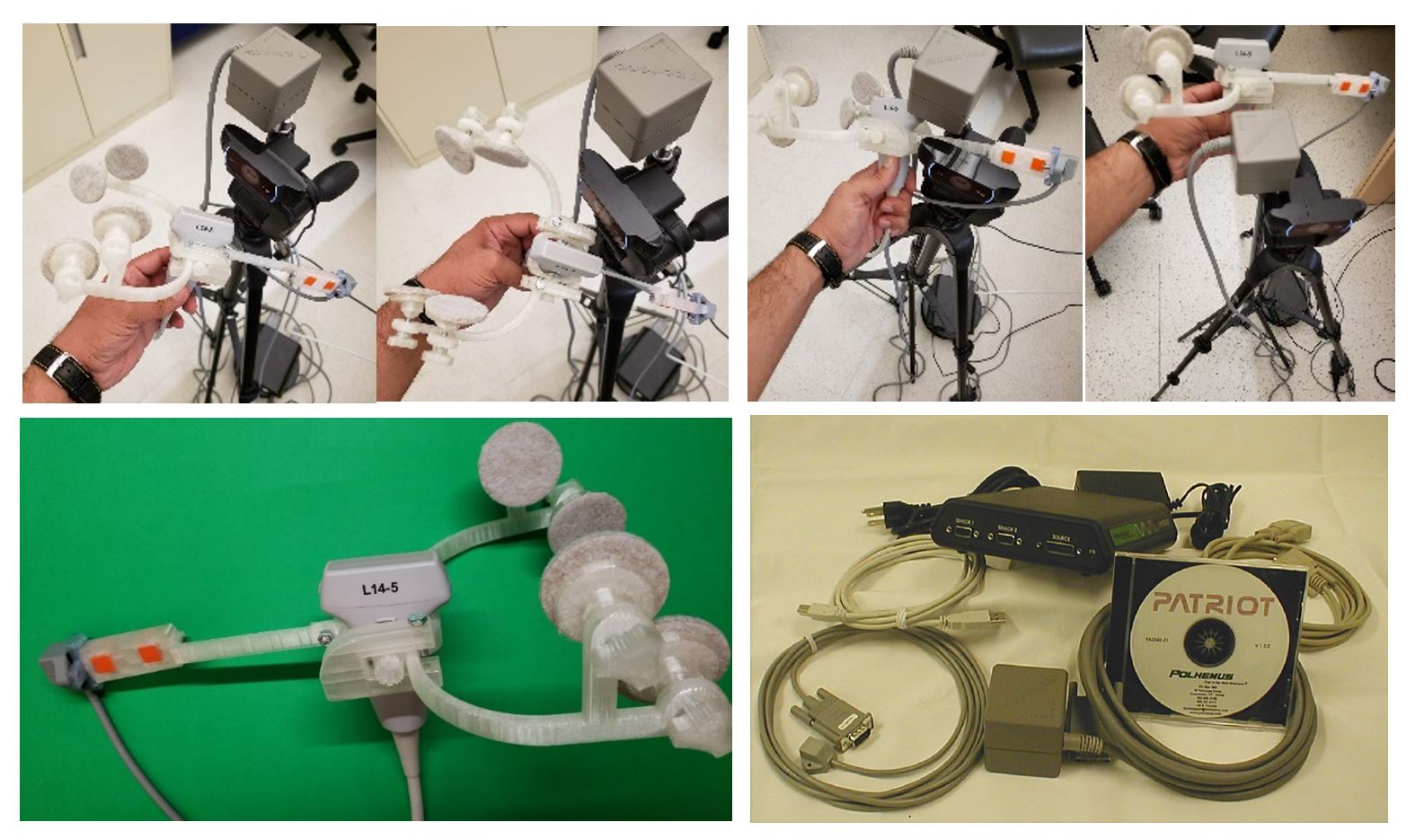}
\caption{\label{fig:FIG10}{First row: different views of magnetic tracking sensors attached on UltraChin. Second row: Different parts of UltraChin and magnetic tracking sensor can be seen in figure.}}
\end{figure*}

\begin{figure}[ht!]
\includegraphics[width=\columnwidth]{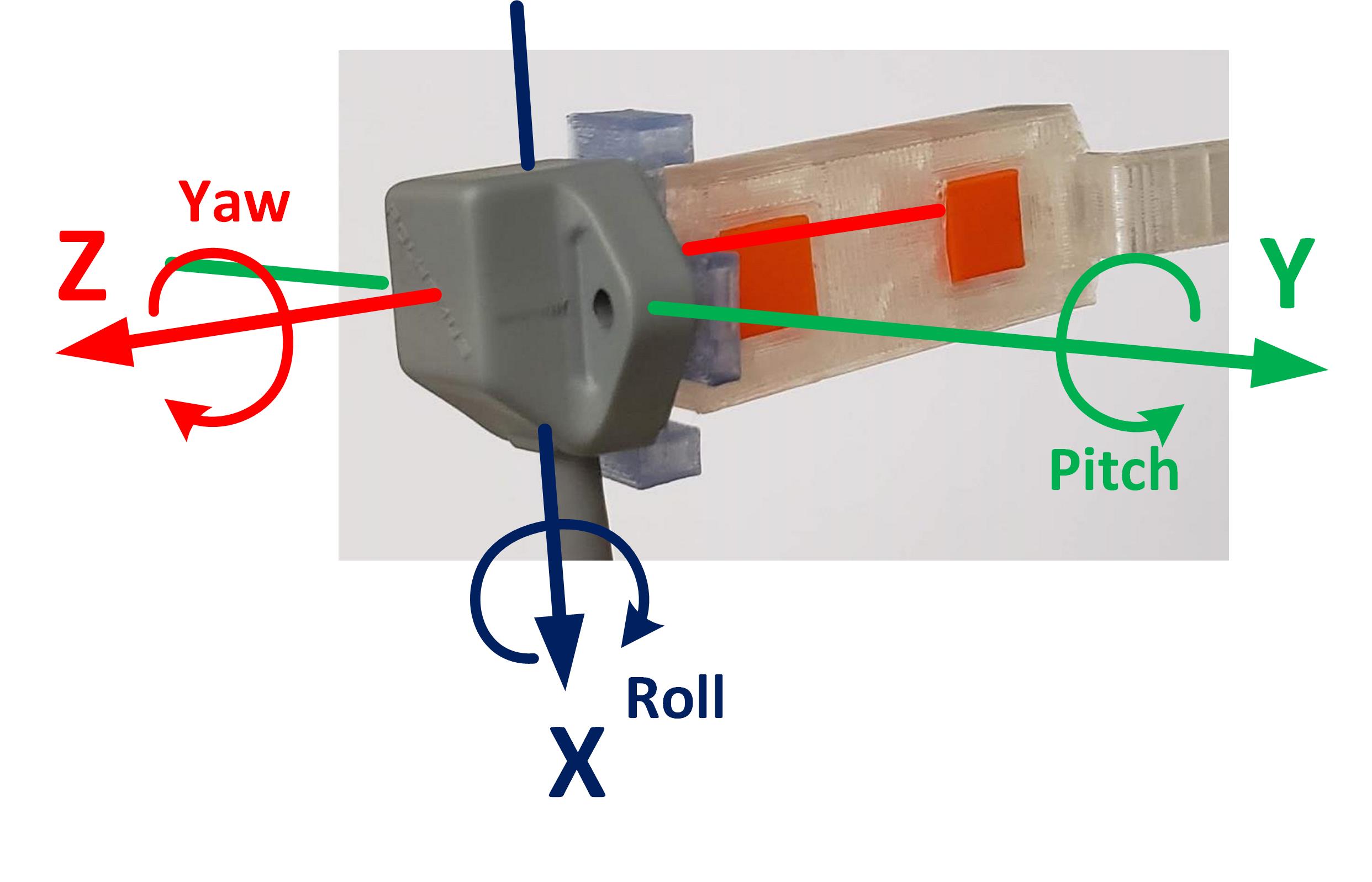}
\caption{\label{fig:FIG11}{Tracking magnetic sensor was attached on UltraChin. Arrows show 6 degree of freedoms defined in our experiments.}}
\end{figure}

Head and probe stabilization is not necessary if the system is only utilized as a pronunciation bio-feedback \cite{gick2005techniques}. However, the accuracy of our system could be improved by adding 3D printable extensions to the UltraChin for head stabilization for a particular linguistic study, while articulatory data will be subject to detailed quantification and measurement. In order to evaluate our 3D printable design, we followed the method in \cite{scobbie2008head}. We attached one magnetic tracking sensors, PATRIOT Polhemus Company (see Figure \ref{fig:FIG10} and \ref{fig:FIG11}), on the UltraChin and participant's chin in two separate experiments. Six degrees of freedom (see Figure \ref{fig:FIG11}) was recorded after 10 times repeating of a similar experiment. For this experiment, the participant's head was fixed using the method in \cite{menard2011development}. We asked the participant to repeat "ho-mo-Maggie" \cite{scobbie2008head} and to open mouth to the maximum position for 10 times. We calculated deviations of the UltraChin in terms of transnational (in millimeters) and rotational (in degree) slippages. 

Table \ref{table:TABLE1} shows the maximum error of the UltraChin after 10 times experiment. For a better understanding of the UltraChin performance, we checked two different settings where four screws of the device were loose (most comfort) or tight (dis-comfort).  
\begin{table*}[]
\center
\small
\begin{tabular}{|c|c|c|c|c|c|c|}
\hline
\multirow{2}{*}{Status of four screws} & \multicolumn{3}{c|}{Max translational in milimeters} & \multicolumn{3}{c|}{Max Rotational in degree} \\ \cline{2-7} 
                          & x           & y           & z          & roll       & yaw       & pitch      \\ \hline
Loose                     & $4.7\pm0.39mm$         & $5.1\pm0.69mm$         & $7.6\pm0.81mm$        & $6.4\pm0.21^{\circ}$        & $4.1\pm0.46^{\circ}$       & $5.9\pm0.86^{\circ}$       \\ \hline
Tight                      & $3.4\pm0.18mm$         & $3.5\pm0.72mm$         & $6.1\pm0.15mm$        & $5.6\pm0.59^{\circ}$        & $3.8\pm0.45^{\circ}$       & $4.7\pm0.91^{\circ}$        \\ \hline
\end{tabular}
\caption{\label{table:TABLE1} Maximum slippage of the UltraChin in 6DOF after 10 time testing on one participant. Values show the mean and standard deviation for each experiment. Screws of the UltraChin was loosely and tightly firm in two different experiments.}
\end{table*}
\\
Our experimental results showed that in the case of tightly firming four screws of UltraChin user's chin has a better long term translational and rotational unwanted slippage without losing a significant comfortability for the user's neck. Slippage errors might be even more due to the usage of cushions, skin deformations, and how the participant is keeping the probe under the chin. In compare to the system in \cite{scobbie2008head}, UltraChin has more long-term slippage in almost all directions. One reason is that UltraChin has fewer stabilizers than previous helmets. Nevertheless, UltraChin errors still are within acceptable deviation limits reported in \cite{scobbie2008head}.
 
\subsection{Evaluation of L2 pronunciation teaching and learning}
\begin{figure*}[ht!]
\centering
\includegraphics[width=0.8\textwidth]{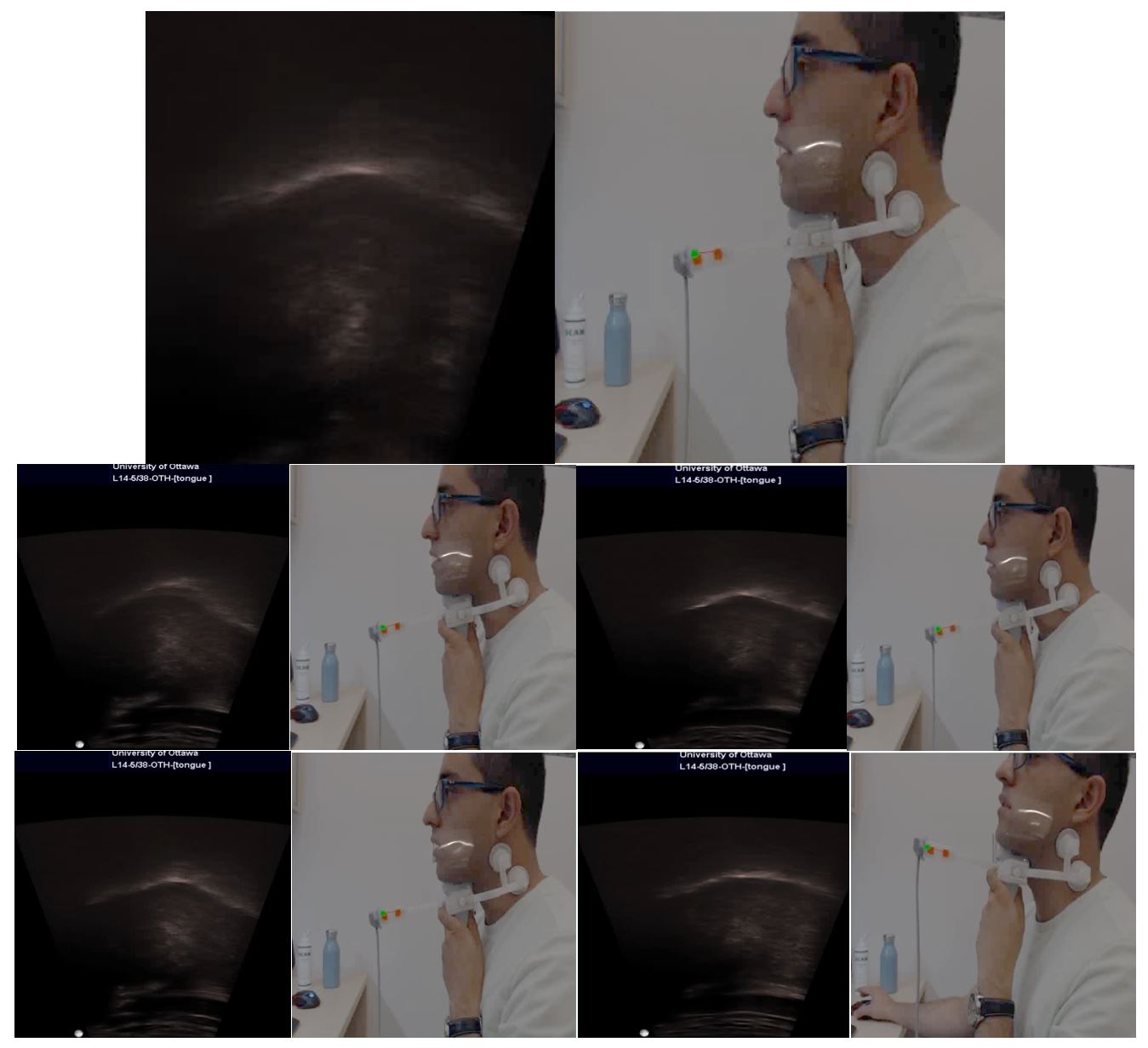}
\caption{\label{fig:FIG12}{Sample frames of our real-time multimodal system. Thick white lines on the right-side images are extracted tongue contours from the corresponding left-side ultrasound images. Red line between two markers is illustrated for each image.}}
\end{figure*}

The positional information from real-time output instances of ProbeNet was used to calculate an estimation of the position, orientation, and scale of ultrasound frames on RGB video frames. In this way, real-time segmented maps from ultrasound frames were predicted by the BowNet model and transferred on face-side of language learner on RGB video frames. Therefore, the language learner can see real-time video frames created by superimposing raw RGB frame, transformed ultrasound image, and tongue segmented image. To illustrate the superimposed image,  we considered different weights for transparency of each image. Figure \ref{fig:FIG12} shows several superimposed samples from our real-time ultrasound-enhanced multimodal system. In the figure, we considered transparency weights as 0.9 for RGB image, 0.4 for Ultrasound image, and 1 for the predicted map, respectively. 
\\
For the sake of representation, we also showed a guideline on the UltraChin to help language learners to keep the probe in a correct position in two-dimensional space (see red lines in Figure \ref{fig:FIG12}). We simultaneously recorded acoustic data, superimposed video, individual video data from two RGB cameras (face-side and front-side view), real-time ultrasound frames, and tongue contour information for later experiments or follow-up study.  

There are many methods and standards in the literature for testing L2 pronunciation acquisition methods \cite{adler2007use, bernhardt2005ultrasound, gick2008ultrasound}. It is possible to use the system for small numbers of L2 pronunciation training individuals \cite{gick2008ultrasound} in a large classroom setting, either by providing individual ultrasound training to language instructors \cite{noguchi2015towards} or by presenting ultrasound videos as part of a blended learning approach \cite{bliss2017using}, or even in a community-based settings \cite{bird2018ultrasound}. For example, in \cite{yamane2015ultrasound}, the previous ultrasound-enhanced system has been tested in several courses at UBC. This kind of systems is also tested for training and revitalization of different indigenous languages \cite{bliss2016ultrasound}. Usability of ultrasound bio-feedback in L2 pronunciation training has been comprehensively investigated in \cite{antolik2019effectiveness, bernhardt2005ultrasound}.
\\
Due to the lack of assessment facilities like a big classroom equipped with ultrasound machines, we followed the method in \cite{gick2008ultrasound} for a participant experiment. In this technique, one approach is used repeatedly to measure the dependent variables from an individual. The dependent variables in this methodology consist of targets to be learned, such as vowels, consonants, or suprasegmentals \cite{gick2008ultrasound}. The main goal is to study articulator positions and segments, the accuracy of production, and speech intelligibility. Good candidates for ultrasound biofeedback are usually vowels, rhotic sounds, retroflex, velar and uvular consonants, and dynamic movements between tongue gestures \cite{bird2018ultrasound}. 
\\
In subsequent studies, ultrasound was used to teach individual challenging sound, such as English /r/, in clinical settings \cite{adler2007use, tateishi2013does}. For this reason, we selected one individual participant to practice predefined letters individually by comparing with the pronunciation of the same statements by a native speaker. 
\\
We utilized sample testing videos from enunciating website UBC language department \cite{yuen2011enunciate} as our truth pronunciation references. An Iranian L2 pronunciation learner volunteered to use our multimodal pronunciation system for ten sessions to improve pronunciation of /r/ sound. Each session contained 20 times repeating of /ri/, /ra/, and /ru/ and comparing with the video downloaded from \cite{yuen2011enunciate}. Before the first session, we trained the participant for correct using our system and watching several training videos from the same website. The feedback from the language learner in the final session was outstanding. We investigated the participant’s awareness of controlling tongue body gestures. Results showed that we gained in performance after a short training session, and this suggests that providing visual feedback, even a short one, improves tongue gesture awareness. 
\\
Benefits of our pronunciation system are not limited to only real-time and automatic characteristics. During pronunciation session, unlike other studies \cite{hueber2013ultraspeech} there is no need any manual synchronization. Furthermore, ultrasound frames, RGB video frames, audio data, overlaid images, and extracted contour information are recorded and visualized simultaneously. Our preliminary assessments showed that a language learner would fatigue slower than previous studies, which the average time was 20 to 30 mins due to maintaining a relatively constant position \cite{gick2008ultrasound}. In our system, non-physical restrictions such as using uniform backgrounds in video recording have been addressed, and the system can be used in any room with different ambient properties. For quantitative study between tongue contours, users can compare tongue contour data in real-time using several statistics which is provided in our system.
\\
Testing the efficacy of our real-time automatic multimodal pronunciation system in details remains in the early stages, and further research should be accomplished to create a fuller and more accurate assessment of our system with the collaboration of linguistics departments. 

\section{Discussion and Conclusion}
\label{section:SEC4}
In this study, we proposed and implemented a fully automatic and real-time modular ultrasound-enhanced multimodal pronunciation training system using several novel innovations. Unlike previous studies, instead of tracking the user’s face or using tracking devices, the ultrasound probe position and orientation are estimated automatically using a 3D printable stabilizer (named UltraChin) and a deep learning model (named ProbeNet). ProbeNet was trained in advance on our dataset to track two markers on the UltraChin. This approach enables our pronunciation system to determine the optimum transformation quantities for multimodal superimposition as a user independence system. 
\\
UltraChin makes the system universal for any type of ultrasound probe as well as invariant respect to the probe image occlusion. UltraChin errors due to the slippage of the device during a speech was within the standard range in the literature. At the same time, pre-trained BowNet model \cite{mozaffari2019bownet}, another deep learning model tracks, delineates and highlights the tongue regions on ultrasound data. Different enhanced and transformed video frames from our different system’s modules are overlaid for illustration in the visualization module. Except for preparation of training datasets, all modules in our system work automatically, in real-time, end-to-end, and without any human manipulation. 
\\
Our preliminary experimental results revealed the significance of our system for better L2 pronunciation training. However, besides all successful performance achievements, the development of our system is still in the early stages. An extensive pedagogical investigation of our pronunciation training system for teaching and learning should be accomplished to evaluate the effectiveness of our system in different aspects of pronunciation training.
\\
Application of our system can even be studied as visual biofeedback (VBF) for other applications like pronunciation learning in different languages or development speech disorders (SSDs) which is a common communication impairment in childhood who consistently exhibit difficulties in the production of specific speech sounds in their native language \cite{ribeiro2019ultrasound}. We believe that publishing our datasets, annotation package, deep learning architectures, and pronunciation training toolkit deployed on a publicly available Python programming language with an easy to use documentation will help other researchers in the different fields of linguistics. Providing a comprehensive GUI for our system is still under progress.

\bibliographystyle{unsrt}
\bibliography{references}
\nocite{*}

\end{document}